\documentclass{article}

\PassOptionsToPackage{numbers, compress}{natbib}
\usepackage[final]{neurips_2025_ml4ps}

\usepackage[utf8]{inputenc} 
\usepackage[T1]{fontenc}    
\usepackage{hyperref}       
\usepackage{url}            
\usepackage{booktabs}       
\usepackage{amsfonts}       
\usepackage{amsmath}
\usepackage{nicefrac}       
\usepackage{microtype}      
\usepackage{xcolor}         
\usepackage{graphicx}

\renewcommand{\P}{\mathbb{P}}
\newcommand{\R}{\mathbb{R}} 

\title{Uncertainty Quantification for Reduced-Order Surrogate Models Applied to Cloud Microphysics}

\author{%
  Jonas E.~Katona
    \\
  Department of Applied \& Computational Mathematics\\
  Yale University\\
  New Haven, CT 06511 \\
  \texttt{jonas.katona@yale.edu} \\
  \And
  Emily K.~de~Jong \\
  Atmospheric, Earth, \& Energy Division \\
  Lawrence Livermore National Laboratory \\
  Livermore, CA 94550 \\
  \texttt{dejong5@llnl.gov} \\
  \And
  Nipun~Gunawardena \\
  Atmospheric, Earth, \& Energy Division \\
  Lawrence Livermore National Laboratory \\
  Livermore, CA 94550 \\
  \texttt{gunawardena1@llnl.gov} \\
}

\begin{document}

\maketitle

\begin{abstract}
Reduced-order models (ROMs) can efficiently simulate high-dimensional physical systems but lack robust uncertainty quantification methods. Existing approaches are frequently architecture- or training-specific, which limits flexibility and generalization. We introduce a post hoc, model-agnostic framework for predictive uncertainty quantification in latent-space ROMs that requires no modification to the underlying architecture or training procedure. Using conformal prediction, our approach estimates statistical prediction intervals for multiple components of the ROM pipeline: latent dynamics, reconstruction, and end-to-end predictions. We demonstrate the method on a latent-space dynamical model for cloud microphysics, where it accurately predicts the evolution of droplet-size distributions and quantifies uncertainty across the ROM pipeline.
\end{abstract}

\section{Introduction}


Latent-space reduced-order modeling learns a compact representation of high-dimensional physical dynamics in a lower-dimensional \textit{latent space}. These models are valuable for scientific applications where the governing physics are partially known or computationally prohibitive to resolve. For example, accurately resolving clouds and precipitation in an atmospheric simulation would require tracking high-dimensional droplet-size distributions (DSDs), a longstanding parametric challenge in climate and weather modeling known as ``cloud microphysics.'' Error and uncertainty in microphysics parameterizations is typically not quantified, but is believed to be a dominant source of uncertainty in future climate projections \cite{morrison_confronting_2020}.

Latent-space ROMs have proven effective in efficiently simulating related complex fluid mechanical systems (e.g. \cite{choi_sns_2020, fries_lasdi_2022}), yet convincing practitioners of their reliability is challenging due to the lack of unified and robust uncertainty quantification (UQ) frameworks. Existing UQ methods for latent-space dynamical models are often tied to specific architectures \cite{Bonnet2023,cheng2025latentspaceenergybasedneural,YONG2025117638}, require expensive training \cite{Simpson2024-ie}, or make parametric assumptions \cite{iakovlev2023latentneuralodessparse,Kingma_2019}.

We present a model-agnostic, post hoc framework for predictive UQ in latent-space ROMs that quantifies uncertainty on \textit{reconstruction}, \textit{latent dynamics}, and \textit{end-to-end predictions} without altering the base architecture or training procedure. Our approach utilizes conformal prediction (CP), a distribution-free method that produces statistically valid prediction intervals---a first for latent-space ROMs. We demonstrate this UQ pipeline on a cloud microphysics ROM trained to predict the evolution of cloud DSDs during coalescence and the formation of precipitation \cite{dejong2025data}, producing reliable UQ estimates that allow practitioners to rigorously evaluate individual components of the ROM architecture.


\section{Proposed UQ framework}

\begin{figure}
    \centering
    \includegraphics[width=\linewidth,trim={0.2cm 1.2cm 2.9cm 0.3cm},clip]{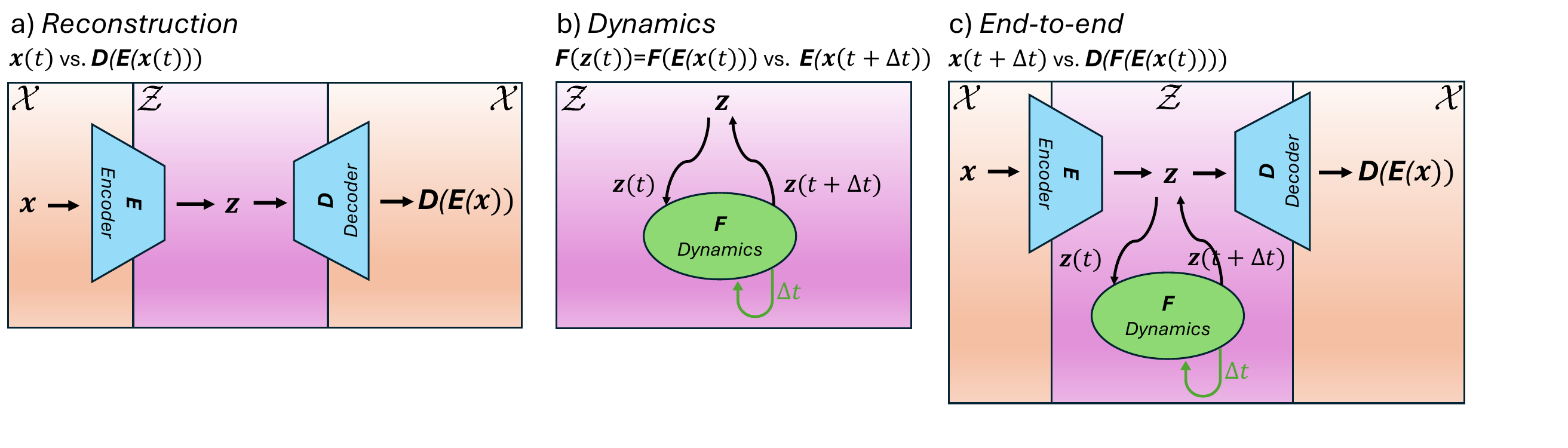}
    \caption{A generic latent-space dynamical model with fixed dynamical time-step $\Delta t$. Figures 1a and 1b show the reconstruction and dynamics sub-models, respectively, that comprise the end-to-end model architecture shown in Figure 1c.}
    \label{fig:latent_dyn}
\end{figure}

A latent-space dynamical ROM consists of a data space, $\mathcal{X}\subseteq\R^{d}$, and a latent space, $\mathcal{Z}\subseteq\R^{m}$, where $m\ll d$; an encoder $E:\mathcal{X}\rightarrow\mathcal{Z}$ and a decoder $D:\mathcal{Z}\rightarrow\mathcal{X}$; and a dynamical system model $F:\mathcal{T}\times\mathcal{Z}\rightarrow\mathcal{Z}$ defined on the latent space, where $\mathcal{T}\subseteq\left[0,T\right]$ for some final time $T>0$.
(See Figure \ref{fig:latent_dyn}.)

We consider the setting where we observe $n$ independent time-evolving realizations of a physical system in $\mathcal{X}$: $\{x_{t}^{(j)}\}_{t\in\mathcal{T}}$ for $j=1,\dots,n$. The proposed pipeline applies regardless of whether model components are trained separately or end-to-end. 
Hence, at \textit{each} fixed time $t\in\mathcal{T}$ and across samples $j=1,\dots,n$, we compute predictive uncertainties in the components of a latent-space dynamical model: 1) reconstruction from the latent space, 2) dynamics in the latent space, and 3) the entire pipeline combined (end-to-end), as shown in Figure \ref{fig:latent_dyn}.

\paragraph{Conformal predictions}

Conformal prediction (CP) provides statistical prediction intervals by computing nonconformity scores on a held-out calibration dataset such that the true outcome $Y$ is contained in the prediction set $\Gamma(X)$ with probability at least $1-\alpha$ \cite{angelopoulos2021gentle,barber_predictive_2021,steinberger_leave-one-out_2016,vovk_cross-conformal_2015,vovk_algorithmic_2022}, i.e., 
\(
\P\left(Y \subseteq \Gamma(X)\right) \geq 1-\alpha
\). This guarantee holds under the assumption of exchangeability of the calibration and test data, satisfied in the case of our DSD dataset due to independence of the sampled initial conditions---see Appendix \ref{data} for details. Because CP requires no changes to model architecture or changing or parametric assumptions on the data distribution, it can be applied either to full outputs or component-wise in multi-dimensional settings to obtain reliable, distribution-free guarantees on predictive coverage \cite{angelopoulos2021gentle}.
We illustrate three common variants of conformal predictions: \textbf{Vanilla conformal} (train--test split, using training data for scoring discrepancies), \textbf{split conformal} (train--validation--test split, scoring on validation set), and \textbf{CV+ conformal} ($k$-fold train--validation splits with aggregated residuals) \cite{angelopoulos2021gentle,barber_predictive_2021,9835501}. CV+ generally yields tighter intervals while maintaining coverage guarantees, with the choice of folds $k$ balancing statistical efficiency and computational overhead \cite{barber_predictive_2021}. 

For DSD-valued predictions, we construct two-sided conformal prediction intervals using the $\alpha/2$ and $1-\alpha/2$ empirical quantiles of the \textit{signed} residuals \cite{barber_predictive_2021}. This approach permits asymmetric upper and lower bounds when the residual distribution is skewed---a crucial feature for DSDs, as these are non-negative and often attain values near zero, allowing prediction intervals to reflect physical uncertainty better. Meanwhile, for latent-space predictions, whose outputs are multivariate and correlated, we instead use a scalar conformity score based on the Mahalanobis distance between predicted and true latent variables, providing a covariance-aware measure of predictive error. In either case, conformal sets are computed at each timestep separately, so that uncertainty can be tracked in time alongside latent-space dynamics and DSD evolution. See Appendix~\ref{tailwise} for further details.

\section{Application to Cloud Microphysics ROM}\label{applications}

We demonstrate the UQ pipeline with a cloud microphysics ROM, trained on DSDs from particle-based simulations of warm-rain coalescence, and developed in detail in \cite{dejong2025data}. Coalescence is the process of small cloud droplets colliding and merging to form larger ones, eventually leading to large rain drops that precipitate \cite{khain_representation_2015}. As shown in Figure \ref{fig:plots}, droplet mass shifts from smaller to larger size bins under coalescence, and the strong nonlinearity of the process drives the emergence and disappearance of multiple modes. 

New parameterizations of cloud microphysics must not only outperform traditional schemes in accuracy and efficiency, but also quantify structural errors and parametric uncertainties that currently hinder the accuracy of large-scale models. Nonlinear ROMs offer a more flexible alternative to bulk parameterizations, which impose restrictive modeling assumptions \cite{kessler_distribution_1969, khairoutdinov_new_2000, seifert_double-moment_2001}, and to linear latent-space ROMs, which could require inefficiently high-dimensional representations to capture DSD coalescence dynamics accurately \cite{lin_enhanced_2025, romor_non-linear_2023, wan_evolve_2023}. Hence, to model droplet coalescence, we combine an autoencoder (AE) for nonlinear dimensionality reduction with parsimonious latent-space ODEs, based on the Sparse Identification of Nonlinear Dynamics (SINDy) technique \cite{brunton_discovering_2016, champion_data-driven_2019,dejong2025data}. 

For details on the AE–SINDy architecture, training, and accuracy, see Appendix \ref{reproducibility}, the associated code repository, and the companion paper by De Jong et al. \cite{dejong2025data}, which develops, trains, and evaluates the AE-SINDy surrogate in depth. The companion paper also applies the UQ framework introduced here to two additional latent-space ROMs---each using a similar autoencoder architecture but different dynamical models in the latent space---thereby highlighting broader context and applicability. Here, we focus specifically on the UQ pipeline and its expanded application to the AE–SINDy surrogate.

\subsection{Dataset from LES with superdroplet method}

We represent each DSD as a mass-density function $\frac{dm}{d\ln r}$, where $r$ is the droplet radius, so that the DSD is defined per logarithmic radius interval. To train and evaluate the AE-SINDy surrogate, we use PSD trajectories generated from large eddy simulations (LES) of warm-rain coalescence with a high-fidelity Lagrangian particle representation of cloud droplets \cite{dejong2025data,shima_super-droplet_2009, unterstrasser_collisional_2020}. DSDs are averaged over a ($200$m)$^3$ cubic domain, filtered for the presence of cloud condensate, and discretized into $N_\text{bins} = 64$ bins uniformly spaced in $\ln r$. For each simulation grid cell, a $600$s coalescence-only forward integration produces the evolution of its PSD from $t = 0$ to $t = 600$s at $\Delta t=10$s intervals. These trajectories provide both the instantaneous DSDs and the associated time derivatives used for training the AE-SINDy model and for assessing predictive uncertainty across different initial conditions.

\subsection{Computational efficiency and UQ}

The LES with Lagrangian microphysics used to generate the dataset are computationally intensive, using $3\times 10^6$ grid cells, each with 128 particles for a total of $\sim 10^8$ Lagrangian particles per simulation. Collisional-coalescence is computed via linear stochastic pairwise sampling within each grid cell \cite{shima_super-droplet_2009}, which scales linearly with the number of particles. Combined with Lagrangian advection and Eulerian-Lagrangian coupling, these simulations tend to require hundreds of CPU-hours for a single multi-hour LES. Traditional Eulerian binned microphysics can be far less expensive, but still evolves $30$--$100$ prognostic DSD bins per cell with quadratically scaling computations for collisional coalescence \cite{khain_representation_2015}.

In contrast, the AE-SINDy surrogate used here compresses each $64$-bin PSD into a $4$-dimensional latent state governed by an ODE system whose evaluation is $\mathcal{O}(1)$ per grid cell per timestep. This yields reductions of several orders of magnitude in computational cost relative to SDM and at least an order of magnitude reduction relative to bin microphysics. 

\begin{figure}
    \centering
    \includegraphics[width=0.98\linewidth,trim={0.3cm 0.3cm 0.2cm 0.2cm},clip]{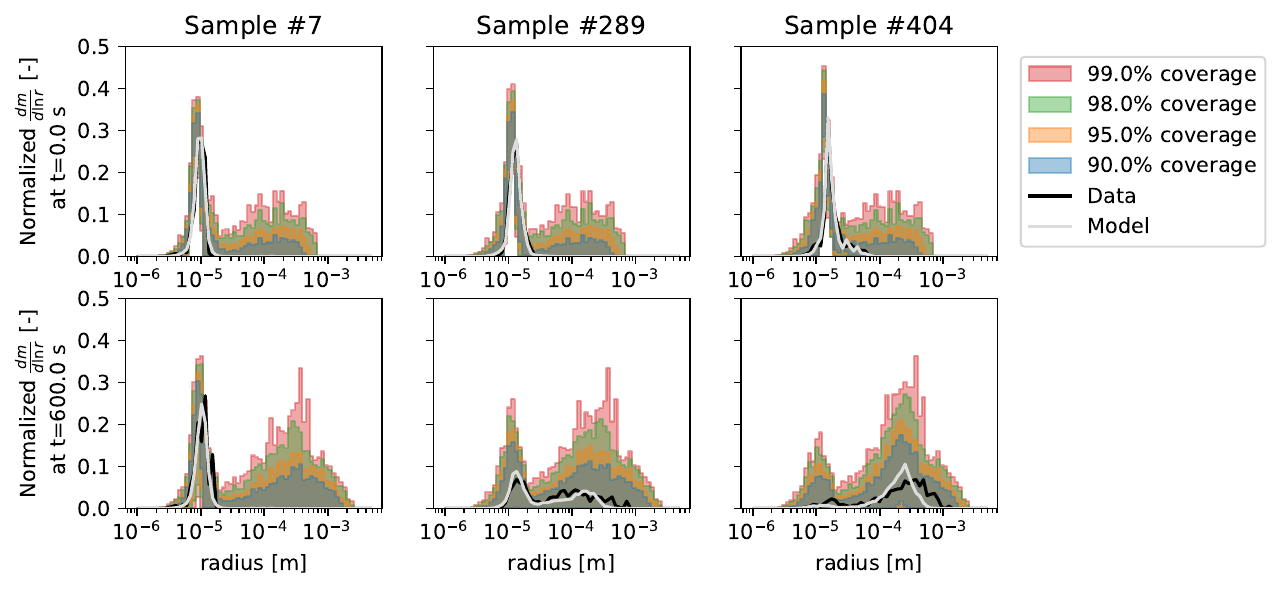}
    \caption{Initial and final states for three sample DSD trajectories from the dataset, predicted using the trained AE-SINDy architecture (``Model'') with empirical prediction intervals provided via CV+ conformal predictions (using $k=20$ folds) at varying nominal coverage levels. These predictions are also compared with the actual DSD final states from the dataset (``Data'').}
    \label{fig:plots}
\end{figure}

\begin{figure}
    \centering
    \includegraphics[width=0.9\linewidth,trim={0.2cm 0.3cm 0.2cm 0.2cm},clip]{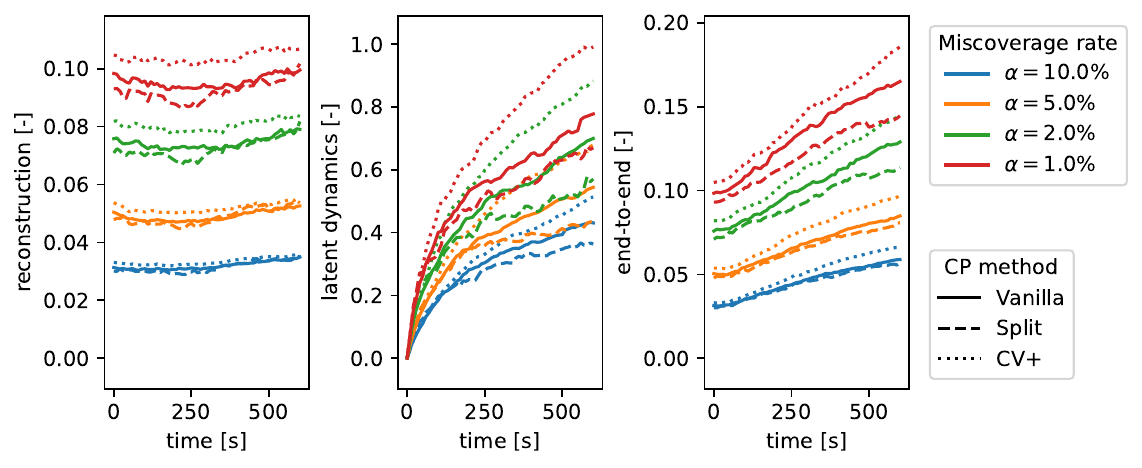}
    \caption{Average prediction interval width vs. time, computed by applying vanilla CP, split CP (using a $60$-$20$-$20$ train-validation-test split), and CV+ conformal predictions (using $k=20$ folds), respectively, to the indicated components of the AE-SINDy pipeline at varying nominal coverage levels $1-\alpha$. Interval widths are computed using the mass-normalized integral of the area between prediction bands across bins for DSD-valued predictions and the volume of the Mahalanobis distance-derived prediction ellipsoids for latent-space predictions.}
    \label{fig:full_errors}
\end{figure}

These reductions motivate the need for rigorous UQ. The UQ pipeline introduced here identifies when the surrogate faithfully reproduces the high-fidelity Lagrangian microphysics and when surrogate uncertainty becomes dominant. Furthermore, this pipeline allows us to measure how both structural uncertainty in the AE-based compression of DSDs and parametric uncertainty in the SINDy-identified latent dynamics propagate through DSD coalescence predictions. 
The examples in Figure~\ref{fig:plots} illustrate prediction intervals for three representative DSDs at fixed times, while Figure~\ref{fig:full_errors} summarizes the uncertainty propagated across different components of the ROM as a function of time and nominal miscoverage level $\alpha$. 

In Figure~\ref{fig:full_errors}, the 
$y$-values report an average prediction interval width for each indicated ROM component. For DSD-valued predictions, this is computed as a normalized, total mass-weighted integral of prediction interval widths across bins, while for latent-space predictions, it is computed via the Mahalanobis distance-derived prediction ellipsoids in latent space, whose geometry is normalized by the residual covariance. Our analysis highlights the ability of the presented UQ pipeline to characterize data-driven ROMs by identifying the specific scales and processes where model improvements are most necessary.

\section{Results \& Discussion}
Figure \ref{fig:plots} shows how predictive uncertainty, as estimated on the testing data using CP, evolves across droplet-size bins during warm-rain coalescence. While end-to-end predictive uncertainty increases with time (cf. Figure \ref{fig:full_errors}), the uncertainty systematically shifts from smaller to larger droplet sizes: the prediction interval ``peak'' at the sub-50$\mu$m cloud-droplet scale tends to contract while the prediction intervals expand markedly at larger rain-droplet scales as coalescence proceeds. This trend even holds for unimodal cloud droplet populations with negligible collisional growth (e.g., sample 7)---nevertheless, uncertainty in the larger rain bins grows with time. Physically, this reflects the inherent difficulty of predicting the onset of rain formation (i.e., the emergence of a secondary right-hand peak), compared to the more stationary evolution of smaller cloud droplets. The result underscores both the interpretability of uncertainty estimates in this framework and a key limitation of this particular ROM: intervals remain widest where precipitation processes begin, highlighting a persistent challenge in modeling warm-rain initiation \cite{morrison_confronting_2020}.

Figure \ref{fig:full_errors} illustrates not only predictive accuracy on unseen data but also qualitative patterns of uncertainty propagation across different components of the AE-SINDy architecture. As anticipated, reconstruction uncertainty intervals, which characterize the autoencoder alone, are consistent across time. The latent dynamics exhibit rapid incipient growth in uncertainty that slows over time, reflecting the challenging cloud-to-rain transition before droplets settle into a rain-dominant coalesced state. By contrast, the full ROM produces nearly linear growth in the end-to-end predictive errors. This highlights a key advantage of component-wise uncertainty analysis in ROMs: we detected how latent errors are effectively ``smoothed out'' by the autoencoder, yielding linear error growth in the final predictions.

Although a dynamical system may evolve on a lower-dimensional manifold, an inaccurate mapping between physical and latent space hinders a faithful, parsimonious latent representation, complicating the modeling of latent dynamics--- particularly in the context of SINDy \cite{brunton_discovering_2016,champion_data-driven_2019}. Figure \ref{fig:full_errors} shows that predicted reconstruction errors remain consistent over time for most of the data, regardless of $\alpha$ or the CP method. Even though predicted errors in the latent dynamics increasingly affect the end-to-end model output accuracy as time progresses, the propagation of these dynamics errors is ultimately mediated by reconstruction back to physical space. Thus, for this cloud microphysics ROM, future work to reduce structural uncertainty in the autoencoder will likely have a greater impact on overall model performance than refining the dynamical model.

While all conformal prediction methods achieved near-nominal coverage (see Appendix \ref{validate}), CV+ yielded wider average intervals to produce more reliable coverage, but at a higher computational cost. Although this can be done in parallel, CV+ requires retraining the surrogate model $k$ times, compared to just once for vanilla or split conformal. The cost of training the surrogate model will therefore determine practical choices for CP techniques in future applications.

\section{Limitations \& Future Directions}

The agreement across CP methods and variability in empirical coverages was notably better for the reconstruction and end-to-end network predictions than for latent dynamical predictions (cf. Figure \ref{fig:full_errors}). This is likely due to greater \textit{variance} in prediction fidelity across times and variables---as well as a wider range of magnitudes overall---in latent predictions compared to normalized DSD predictions. Increasing training data---for instance, by altering the initial conditions or dynamical driver of the cloud LES (see Appendix \ref{data} or \cite{dejong2025data})---could reduce this variability, yielding more consistent CP intervals and improving coverage accuracy on the test set.

While this work demonstrates the flexibility of conformal prediction for uncertainty quantification in general black-box architectures, a key limitation of standard CP methods is that prediction intervals are scaled only \textit{relative} to the input variables; the \textit{width} of a given interval at a particular output and time remains fixed across the dataset. Although adaptive variants can adjust interval widths to reflect varying uncertainty \cite{angelopoulos2021gentle,barber2022limits,colombo2023training,guan2023localized,lei2018distribution,romano2019conformalized}, we do not explore these extensions in this study.

That being said, the post hoc UQ approach introduced in this work is not limited to conformal prediction and could also extend to other interval- and set-valued UQ methods---e.g., parametric prediction intervals, confidence intervals, and Bayesian credible intervals \cite{casella_statistical_2002,gelman_bayesian_2013,hyndman_forecasting_2021,montgomery_applied_2014,wasserman_all_2004}. Exploring these extensions, especially on other ROMs, could show the usefulness of this approach for quantifying other types of uncertainty in surrogate modeling pipelines.

Taken together with the companion study De Jong et al. \cite{dejong2025data}, which develops the AE-SINDy surrogate itself, this work lays the foundation for a unified framework for both rigorous UQ of latent-space ROMs and its relevance in the efficient modeling of warm-rain microphysics.

\begin{ack}
This work was performed in part under the auspices of the U.S. Department of Energy by Lawrence Livermore National Laboratory (LLNL) under Contract DE-AC52-07NA27344 and supported by the Laboratory Directed Research and Development Program (LDRD), project number 25-ERD-045. J. Katona was partially supported by a stipend and teaching fellowship from the Yale Graduate School of Arts and Sciences. The authors have declared that none of them have any competing interests. Released under IM number LLNL-CONF-2010541. 

The authors would like to thank LLNL---particularly Livermore Computing---for their support and for providing high-performance computing resources that enabled the simulations, data analyses, and figure generation in this study. The authors also appreciate the support and valuable advice of Peter Caldwell, Hassan Beydoun, Aaron Donahue, Chris Golaz, Youngsoo Choi, and many others in AEED and CASC at LLNL. 
\end{ack}

\bibliography{UQ}

@article{morrison_confronting_2020,
	title = {Confronting the Challenge of Modeling Cloud and Precipitation Microphysics},
	volume = {12},
	issn = {1942-2466},
	doi = {10.1029/2019MS001689},
	number = {8},
	journal = {Journal of Advances in Modeling Earth Systems},
	author = {Morrison, Hugh and Lier‐Walqui, Marcus van and Fridlind, Ann M. and Grabowski, Wojciech W. and Harrington, Jerry Y. and Hoose, Corinna and Korolev, Alexei and Kumjian, Matthew R. and Milbrandt, Jason A. and Pawlowska, Hanna and Posselt, Derek J. and Prat, Olivier P. and Reimel, Karly J. and Shima, Shin-Ichiro and Diedenhoven, Bastiaan van and Xue, Lulin},
	year = {2020},
	langid = {english},
}

@article{shima_super-droplet_2009,
	title = {The super-droplet method for the numerical simulation of clouds and precipitation: a particle-based and probabilistic microphysics model coupled with a non-hydrostatic model},
	volume = {135},
	issn = {1477-870X},
	doi = {10.1002/qj.441},
	shorttitle = {The super-droplet method for the numerical simulation of clouds and precipitation},
	pages = {1307--1320},
	number = {642},
	journal = {Quarterly Journal of the Royal Meteorological Society},
	author = {Shima, S. and Kusano, K. and Kawano, A. and Sugiyama, T. and Kawahara, S.},
	year = {2009},
	langid = {english},
}

@article{champion_data-driven_2019,
	title = {Data-driven discovery of coordinates and governing equations},
	volume = {116},
	url = {https://www.pnas.org/doi/10.1073/pnas.1906995116},
	doi = {10.1073/pnas.1906995116},
	pages = {22445--22451},
	number = {45},
	journal = {Proceedings of the National Academy of Sciences},
	author = {Champion, Kathleen and Lusch, Bethany and Kutz, J. Nathan and Brunton, Steven L.},
	urldate = {2024-05-28},
	date = {2019-11-05},
year = {2019},
}

@article{brunton_discovering_2016,
	title = {Discovering governing equations from data by sparse identification of nonlinear dynamical systems},
	volume = {113},
	url = {https://www.pnas.org/doi/10.1073/pnas.1517384113},
	doi = {10.1073/pnas.1517384113},
	pages = {3932--3937},
	number = {15},
	journal = {Proceedings of the National Academy of Sciences},
	author = {Brunton, Steven L. and Proctor, Joshua L. and Kutz, J. Nathan},
	urldate = {2025-08-21},
	date = {2016-04-12},
        year = {2016},
	note = {Publisher: Proceedings of the National Academy of Sciences},
}

@article{romor_non-linear_2023,
	title = {Non-linear Manifold Reduced-Order Models with Convolutional Autoencoders and Reduced Over-Collocation Method {\textbar} Journal of Scientific Computing},
	volume = {94},
	url = {https://link.springer.com/article/10.1007/s10915-023-02128-2},
	doi = {10.1007/s10915-023-02128-2},
	number = {74},
	journal = {Journal of Scientific Computing},
	author = {Romor, Francesco and Stabile, Giovanni and Rozza, Gianluigi},
	urldate = {2025-08-21},
	year = {2023},
}

@article{lin_enhanced_2025,
	title = {An Enhanced Reduced-Order Model Based on Dynamic Mode Decomposition for Advection-Dominated Problems},
	volume = {102},
	issn = {0885-7474},
	url = {https://doi.org/10.1007/s10915-025-02820-5},
	doi = {10.1007/s10915-025-02820-5},
	number = {3},
	journal = {J. Sci. Comput.},
	author = {Lin, Yifan and Gao, Zhen},
	urldate = {2025-08-21},
	date = {2025-02-05},
year = {2025}
}

@misc{wan_evolve_2023,
	title = {Evolve Smoothly, Fit Consistently: Learning Smooth Latent Dynamics For Advection-Dominated Systems},
	url = {http://arxiv.org/abs/2301.10391},
	doi = {10.48550/arXiv.2301.10391},
	shorttitle = {Evolve Smoothly, Fit Consistently},
	number = {{arXiv}:2301.10391},
	publisher = {{arXiv}},
	author = {Wan, Zhong Yi and Zepeda-Núñez, Leonardo and Boral, Anudhyan and Sha, Fei},
	urldate = {2025-08-21},
	date = {2023-02-06},
        year = {2023},
	eprinttype = {arxiv},
	eprint = {2301.10391 [cs]},
}

@book{vovk_algorithmic_2022,
	location = {Cham},
	title = {Algorithmic Learning in a Random World},
	rights = {https://www.springernature.com/gp/researchers/text-and-data-mining},
	isbn = {978-3-031-06648-1 978-3-031-06649-8},
	url = {https://link.springer.com/10.1007/978-3-031-06649-8},
	publisher = {Springer International Publishing},
	author = {Vovk, Vladimir and Gammerman, Alexander and Shafer, Glenn},
	urldate = {2025-08-21},
	year = {2022},
	langid = {english},
	doi = {10.1007/978-3-031-06649-8},
}

@misc{steinberger_leave-one-out_2016,
	title = {Leave-one-out prediction intervals in linear regression models with many variables},
	url = {http://arxiv.org/abs/1602.05801},
	doi = {10.48550/arXiv.1602.05801},
	number = {{arXiv}:1602.05801},
	publisher = {{arXiv}},
	author = {Steinberger, Lukas and Leeb, Hannes},
	urldate = {2025-08-21},
	date = {2016-02-18},
year = {2016},
	eprinttype = {arxiv},
	eprint = {1602.05801 [math]},
}

@article{barber_predictive_2021,
	title = {Predictive inference with the jackknife+},
	volume = {49},
	issn = {0090-5364, 2168-8966},
	url = {https://projecteuclid.org/journals/annals-of-statistics/volume-49/issue-1/Predictive-inference-with-the-jackknife/10.1214/20-AOS1965.full},
	doi = {10.1214/20-AOS1965},
	pages = {486--507},
	number = {1},
	journal = {The Annals of Statistics},
	author = {Barber, Rina Foygel and Candès, Emmanuel J. and Ramdas, Aaditya and Tibshirani, Ryan J.},
	urldate = {2025-08-21},
	date = {2021-02},
year = {2021},
	note = {Publisher: Institute of Mathematical Statistics},
}

@article{vovk_cross-conformal_2015,
	title = {Cross-conformal predictors},
	volume = {74},
	issn = {1573-7470},
	url = {https://doi.org/10.1007/s10472-013-9368-4},
	doi = {10.1007/s10472-013-9368-4},
	pages = {9--28},
	number = {1},
	journal = {Annals of Mathematics and Artificial Intelligence},
	shortjournal = {Ann Math Artif Intell},
	author = {Vovk, Vladimir},
	urldate = {2025-08-21},
	date = {2015-06-01},
year = {2015},
	langid = {english},
}

@incollection{vaart1998chapter21,
  author    = {van der Vaart, A. W.},
  title     = {Chapter 21},
  booktitle = {Asymptotic Statistics},
  publisher = {Cambridge University Press},
  year      = {1998},
  doi       = {10.1017/CBO9780511802256},
pages = {304-315}
}

@incollection{serfling1980chapter2,
  author    = {Serfling, Robert J.},
  title     = {Chapter 2},
  booktitle = {Approximation Theorems of Mathematical Statistics},
  publisher = {Wiley},
  year      = {1980},
  doi       = {10.1002/9780470316481},
  pages     = {74-107} 
}

@article{lei2018distribution,
  author  = {Lei, Jing and Rinaldo, Alessandro and Wasserman, Larry},
  title   = {Distribution-Free Predictive Inference for Regression},
  journal = {Journal of the American Statistical Association},
  volume  = {113},
  number  = {523},
  pages   = {1094--1111},
  year    = {2018},
  doi     = {10.1080/01621459.2017.1307116}
}

@inproceedings{romano2019conformalized,
  author    = {Romano, Yaniv and Patterson, Evan and Cand{\`e}s, Emmanuel J.},
  title     = {Conformalized Quantile Regression},
  booktitle = {Advances in Neural Information Processing Systems},
  volume    = {32},
  year      = {2019},
  url       = {https://papers.nips.cc/paper/2019/hash/5103c3584b063c431bd1268e9b5e76fb-Abstract.html}
}

@article{guan2023localized,
  author  = {Guan, Leying and Lei, Jing},
  title   = {Localized Conformal Prediction: A Generalized Framework},
  journal = {Biometrika},
  volume  = {110},
  number  = {1},
  pages   = {33--50},
  year    = {2023},
  doi     = {10.1093/biomet/asac042},
  url     = {https://academic.oup.com/biomet/article/110/1/33/6647831}
}

@inproceedings{colombo2023training,
  author    = {Colombo, Nicol{\`o} and Vovk, Vladimir and Guo, Ruoqi and Lei, Jing and Nouretdinov, Ilia},
  title     = {On Training Locally Adaptive Conformal Predictors},
  booktitle = {Proceedings of the 40th International Conference on Machine Learning},
  series    = {Proceedings of Machine Learning Research},
  volume    = {204},
  pages     = {6235--6258},
  year      = {2023},
  url       = {https://proceedings.mlr.press/v204/colombo23a/colombo23a.pdf}
}

@article{barber2022limits,
  author  = {Barber, Rina Foygel and Cand{\`e}s, Emmanuel J. and Ramdas, Aaditya and Tibshirani, Ryan J.},
  title   = {The Limits of Distribution-Free Conditional Predictive Inference},
  journal = {Proceedings of the National Academy of Sciences},
  volume  = {119},
  number  = {34},
  pages   = {e2204569119},
  year    = {2022},
  doi     = {10.1073/pnas.2204569119},
  url     = {https://www.pnas.org/doi/10.1073/pnas.2204569119}
}

@misc{angelopoulos2021gentle,
  author       = {Angelopoulos, Anastasios N. and Bates, Stephen},
  title        = {A Gentle Introduction to Conformal Prediction and Distribution-Free Uncertainty Quantification},
  year         = {2021},
  howpublished = {arXiv preprint arXiv:2107.07511},
  url          = {https://people.eecs.berkeley.edu/~angelopoulos/publications/downloads/gentle_intro_conformal_dfuq.pdf}
}

@article{fries_lasdi_2022,
	title = {{LaSDI}: Parametric Latent Space Dynamics Identification},
	volume = {399},
	issn = {0045-7825},
	url = {https://www.sciencedirect.com/science/article/pii/S0045782522004807},
	doi = {10.1016/j.cma.2022.115436},
	shorttitle = {{LaSDI}},
	pages = {115436},
	journal = {Computer Methods in Applied Mechanics and Engineering},
	shortjournal = {Computer Methods in Applied Mechanics and Engineering},
	author = {Fries, William D. and He, Xiaolong and Choi, Youngsoo},
	urldate = {2025-08-26},
	date = {2022-09-01},
year = {2022},
}

@misc{choi_sns_2020,
	title = {{SNS}: A Solution-based Nonlinear Subspace method for time-dependent model order reduction},
	url = {http://arxiv.org/abs/1809.04064},
	doi = {10.48550/arXiv.1809.04064},
	shorttitle = {{SNS}},
	number = {{arXiv}:1809.04064},
	publisher = {{arXiv}},
	author = {Choi, Youngsoo and Coombs, Deshawn and Anderson, Robert},
	urldate = {2025-08-26},
	date = {2020-01-02},
year = {2020},
	eprinttype = {arxiv},
	eprint = {1809.04064 [cs]},
}

@article{unterstrasser_collisional_2020,
	title = {Collisional growth in a particle-based cloud microphysical model: insights from column model simulations using {LCM1D} (v1.0)},
	volume = {13},
	issn = {1991-959X},
	shorttitle = {Collisional growth in a particle-based cloud microphysical model},
	url = {https://gmd.copernicus.org/articles/13/5119/2020/},
	doi = {10.5194/gmd-13-5119-2020},
	language = {English},
	number = {11},
	urldate = {2023-01-24},
	journal = {Geoscientific Model Development},
	author = {Unterstrasser, Simon and Hoffmann, Fabian and Lerch, Marion},
	month = oct,
	year = {2020},
	note = {Publisher: Copernicus GmbH},
	pages = {5119--5145},
}

@article{Kingma_2019,
   title={An Introduction to Variational Autoencoders},
   volume={12},
   ISSN={1935-8245},
   url={http://dx.doi.org/10.1561/2200000056},
   DOI={10.1561/2200000056},
   number={4},
   journal={Foundations and Trends® in Machine Learning},
   publisher={Now Publishers},
   author={Kingma, Diederik P. and Welling, Max},
   year={2019},
   pages={307–392} }

@misc{iakovlev2023latentneuralodessparse,
      title={Latent Neural ODEs with Sparse Bayesian Multiple Shooting}, 
      author={Valerii Iakovlev and Cagatay Yildiz and Markus Heinonen and Harri Lähdesmäki},
      year={2023},
      eprint={2210.03466},
      archivePrefix={arXiv},
      primaryClass={cs.LG},
      url={https://arxiv.org/abs/2210.03466}, 
}

@misc{cheng2025latentspaceenergybasedneural,
      title={Latent Space Energy-based Neural ODEs}, 
      author={Sheng Cheng and Deqian Kong and Jianwen Xie and Kookjin Lee and Ying Nian Wu and Yezhou Yang},
      year={2025},
      eprint={2409.03845},
      archivePrefix={arXiv},
      primaryClass={cs.LG},
      url={https://arxiv.org/abs/2409.03845}, 
}

@article{YONG2025117638,
title = {Learning latent space dynamics with model-form uncertainties: A stochastic reduced-order modeling approach},
journal = {Computer Methods in Applied Mechanics and Engineering},
volume = {435},
pages = {117638},
year = {2025},
issn = {0045-7825},
doi = {https://doi.org/10.1016/j.cma.2024.117638},
url = {https://www.sciencedirect.com/science/article/pii/S0045782524008922},
author = {Jin Yi Yong and Rudy Geelen and Johann Guilleminot}
}

@ARTICLE{Simpson2024-ie,
  title     = "{VpROM}: a novel variational autoencoder-boosted reduced order
               model for the treatment of parametric dependencies in nonlinear
               systems",
  author    = "Simpson, Thomas and Vlachas, Konstantinos and Garland, Anthony
               and Dervilis, Nikolaos and Chatzi, Eleni",
  journal   = "Sci. Rep.",
  publisher = "Springer Science and Business Media LLC",
  volume    =  14,
  number    =  1,
  pages     = "6091",
  month     =  mar,
  year      =  2024,
  copyright = "https://creativecommons.org/licenses/by/4.0",
  language  = "en",
doi = "10.1038/s41598-024-56118-x"
}

@article{Bonnet2023,
  title = {Bringing uncertainty quantification to the extreme-edge with memristor-based Bayesian neural networks},
  volume = {14},
  ISSN = {2041-1723},
  url = {http://dx.doi.org/10.1038/s41467-023-43317-9},
  DOI = {10.1038/s41467-023-43317-9},
  number = {1},
  journal = {Nature Communications},
  publisher = {Springer Science and Business Media LLC},
  author = {Bonnet,  Djohan and Hirtzlin,  Tifenn and Majumdar,  Atreya and Dalgaty,  Thomas and Esmanhotto,  Eduardo and Meli,  Valentina and Castellani,  Niccolo and Martin,  Simon and Nodin,  Jean-Fran\c{c}ois and Bourgeois,  Guillaume and Portal,  Jean-Michel and Querlioz,  Damien and Vianello,  Elisa},
  year = {2023},
  month = nov 
}

@article{khain_representation_2015,
	title = {Representation of microphysical processes in cloud-resolving models: Spectral (bin) microphysics versus bulk parameterization},
	volume = {53},
	issn = {1944-9208},
	url = {https://onlinelibrary.wiley.com/doi/abs/10.1002/2014RG000468},
	doi = {10.1002/2014RG000468},
	shorttitle = {Representation of microphysical processes in cloud-resolving models},
	pages = {247--322},
	number = {2},
	journal = {Reviews of Geophysics},
	author = {Khain, A. P. and Beheng, K. D. and Heymsfield, A. and Korolev, A. and Krichak, S. O. and Levin, Z. and Pinsky, M. and Phillips, V. and Prabhakaran, T. and Teller, A. and van den Heever, S. C. and Yano, J.-I.},
	urldate = {2022-08-04},
	year = {2015},
	langid = {english},
	note = {\_eprint: https://onlinelibrary.wiley.com/doi/pdf/10.1002/2014RG000468},
}

@article{khairoutdinov_new_2000,
	title = {A New Cloud Physics Parameterization in a Large-Eddy Simulation Model of Marine Stratocumulus},
	volume = {128},
	issn = {1520-0493, 0027-0644},
	url = {https://journals.ametsoc.org/view/journals/mwre/128/1/1520-0493_2000_128_0229_ancppi_2.0.co_2.xml},
	doi = {10.1175/1520-0493(2000)128<0229:ANCPPI>2.0.CO;2},
	pages = {229--243},
	number = {1},
	journal = {Monthly Weather Review},
	author = {Khairoutdinov, Marat and Kogan, Yefim},
	urldate = {2024-05-21},
	date = {2000-01-01},
year = {2000},
	note = {Publisher: American Meteorological Society
Section: Monthly Weather Review},
}

@article{kessler_distribution_1969,
	title = {On the Distribution and Continuity of Water Substance in Atmospheric Circulations},
	doi = {10.1007/978-1-935704-36-2_1},
	series = {Meteorological Monographs},
	pages = {1--84},
	journal = {On the Distribution and Continuity of Water Substance in Atmospheric Circulations},
	author = {Kessler, Edwin},
	editor = {Kessler, Edwin},
	year = {1969},
	langid = {english},
}

@article{seifert_double-moment_2001,
	title = {A double-moment parameterization for simulating autoconversion, accretion and selfcollection},
	volume = {59-60},
	issn = {0169-8095},
	url = {https://www.sciencedirect.com/science/article/pii/S0169809501001260},
	doi = {10.1016/S0169-8095(01)00126-0},
	series = {13th International Conference on Clouds and Precipitation},
	pages = {265--281},
	journal = {Atmospheric Research},
	shortjournal = {Atmospheric Research},
	author = {Seifert, Axel and Beheng, Klaus D.},
	urldate = {2025-05-06},
	date = {2001-10-01},
year = {2001},
}

@INPROCEEDINGS{9835501,
  author={Thirumuruganathan, Saravanan and Shetiya, Suraj and Koudas, Nick and Das, Gautam},
  booktitle={2022 IEEE 38th International Conference on Data Engineering (ICDE)}, 
  title={Prediction Intervals for Learned Cardinality Estimation: An Experimental Evaluation}, 
  year={2022},
  volume={},
  number={},
  pages={3051-3064},
  doi={10.1109/ICDE53745.2022.00274}}

@book{wasserman_all_2004,
  title        = {All of Statistics: A Concise Course in Statistical Inference},
  author       = {Wasserman, Larry},
  year         = {2004},
  publisher    = {Springer},
  address      = {New York},
  doi          = {10.1007/978-0-387-21736-9},
  isbn         = {9780387402727}
}

@book{casella_statistical_2002,
  title        = {Statistical Inference},
  author       = {Casella, George and Berger, Roger L.},
  edition      = {2},
  year         = {2002},
  publisher    = {Duxbury},
  address      = {Pacific Grove, CA},
  isbn         = {9780534243128}
}

@book{gelman_bayesian_2013,
  title        = {Bayesian Data Analysis},
  author       = {Gelman, Andrew and Carlin, John B. and Stern, Hal S. and Dunson, David B. and Vehtari, Aki and Rubin, Donald B.},
  edition      = {3},
  year         = {2013},
  publisher    = {CRC Press},
  address      = {Boca Raton, FL},
  doi          = {10.1201/b16018},
  isbn         = {9781439840955}
}

@book{montgomery_applied_2014,
  title        = {Applied Statistics and Probability for Engineers},
  author       = {Montgomery, Douglas C. and Runger, George C.},
  edition      = {6},
  year         = {2014},
  publisher    = {Wiley},
  address      = {Hoboken, NJ},
  doi          = {10.1002/9781118762885},
  isbn         = {9781118539715}
}

@book{hyndman_forecasting_2021,
  title        = {Forecasting: Principles and Practice},
  author       = {Hyndman, Rob J. and Athanasopoulos, George},
  edition      = {3},
  year         = {2021},
  publisher    = {OTexts},
  note         = {Available online at \url{https://otexts.com/fpp3/}}
}

@article{LedoitWolf2012,
  author       = {Olivier Ledoit and Michael Wolf},
  title        = {Nonlinear Shrinkage Estimation of Large-Dimensional Covariance Matrices},
  journal      = {Annals of Statistics},
  year         = {2012},
  volume       = {40},
  number       = {2},
  pages        = {1024--1060},
  doi          = {10.1214/12-AOS989},
}

@article{scikit-learn,
    title={Scikit-learn: Machine Learning in {P}ython},
    author={Pedregosa, F. and Varoquaux, G. and Gramfort, A. and Michel, V.
            and Thirion, B. and Grisel, O. and Blondel, M. and Prettenhofer, P.
            and Weiss, R. and Dubourg, V. and Vanderplas, J. and Passos, A. and
            Cournapeau, D. and Brucher, M. and Perrot, M. and Duchesnay, E.},
    journal={Journal of Machine Learning Research},
    volume={12},
    pages={2825--2830},
    year={2011}
   }

@inproceedings{Optuna,
author = {Akiba, Takuya and Sano, Shotaro and Yanase, Toshihiko and Ohta, Takeru and Koyama, Masanori},
title = {Optuna: A Next-generation Hyperparameter Optimization Framework},
year = {2019},
isbn = {9781450362016},
publisher = {Association for Computing Machinery},
address = {New York, NY, USA},
url = {https://doi.org/10.1145/3292500.3330701},
doi = {10.1145/3292500.3330701},
booktitle = {Proceedings of the 25th ACM SIGKDD International Conference on Knowledge Discovery \& Data Mining},
pages = {2623–2631},
location = {Anchorage, AK, USA},
series = {KDD '19}
}

@article{dejong2025data,
  author    = {De Jong, Emily K. and Gunawardena, Nipun and Katona, Jonas Erno and Beydoun, Hassan and Ghosh, Debojyoti and Caldwell, Peter Martin},
  title     = {Data-Driven Reduced Ordering Modeling for Warm Rain Microphysics},
  journal   = {Authorea},
  year      = {2025},
  month     = nov,
  doi       = {10.22541/au.176220214.46858428/v1},
  url       = {https://www.authorea.com/doi/full/10.22541/au.176220214.46858428/v1}
}

\appendix

\section{Validation of empirical prediction intervals}\label{validate}

Tables \ref{tab:cp_coverage_mean_std} and \ref{tab:cp_coverage_medians} give summary statistics---means and standard deviations, and medians, respectively---for the nominal coverages in the outputs for different subsets of the network: autoencoder/reconstruction alone, latent dynamical model alone, and the entire end-to-end network predictions. These statistics are averaged over all output variables---64 bins for DSD data or 4 latent variables---and times---61 timesteps at 10-second intervals.

\begin{table}[ht]
\centering
\small
\caption{Empirical coverage (in \%, given as \textbf{empirical mean $\pm$ empirical standard deviation} across all times and output coordinates) for prediction intervals at marginal coverage levels 90\%, 95\%, 98\%, and 99\%. Split conformal was applied on a 60-20-20 train-validation-test split, and CV+ conformal was applied with $k=20$ folds. Empirical medians appear in Table \ref{tab:cp_coverage_medians}.}
\begin{tabular}{llcccc}
\toprule
Sub-model & CP Method & $1-\alpha=90$\% & $1-\alpha=95$\% & $1-\alpha=98$\% & $1-\alpha=99$\% \\ 
\midrule
Reconstruction & Vanilla & 88.56 $\pm$ 3.16 & 93.86 $\pm$ 2.30 & 96.96 $\pm$ 1.65 & 98.00 $\pm$ 1.38 \\
Reconstruction & Split   & 87.70 $\pm$ 4.31 & 92.87 $\pm$ 3.62 & 96.10 $\pm$ 2.57 & 97.34 $\pm$ 2.17 \\
Reconstruction & CV+     & 89.04 $\pm$ 3.09 & 94.36 $\pm$ 2.21 & 97.34 $\pm$ 1.60 & 98.28 $\pm$ 1.35 \\
\midrule
Latent dynamics & Vanilla & 89.38 $\pm$ 5.41 & 95.16 $\pm$ 3.36 & 97.88 $\pm$ 1.64 & 98.98 $\pm$ 1.04 \\
Latent dynamics & Split   & 88.22 $\pm$ 4.33 & 94.30 $\pm$ 1.87 & 97.32 $\pm$ 1.11 & 98.44 $\pm$ 1.26 \\
Latent dynamics & CV+     & 95.23 $\pm$ 3.45 & 98.36 $\pm$ 1.93 & 99.44 $\pm$ 1.19 & 99.83 $\pm$ 0.53 \\
\midrule
End-to-end & Vanilla & 88.65 $\pm$ 3.35 & 93.73 $\pm$ 2.55 & 96.79 $\pm$ 1.96 & 97.94 $\pm$ 1.54 \\
End-to-end & Split   & 87.36 $\pm$ 4.57 & 92.79 $\pm$ 3.55 & 96.25 $\pm$ 2.54 & 97.45 $\pm$ 1.96 \\
End-to-end & CV+     & 90.56 $\pm$ 3.57 & 95.20 $\pm$ 2.43 & 97.64 $\pm$ 1.68 & 98.46 $\pm$ 1.37 \\
\bottomrule
\end{tabular}
\label{tab:cp_coverage_mean_std}
\end{table}

\begin{table}[ht]
\centering
\caption{Empirical coverage (in \%, given as the \textbf{empirical median} across all times and output coordinates) for prediction intervals at marginal coverage levels $1-\alpha=$90\%, 95\%, 98\%, and 99\%. Split conformal was applied on a 60-20-20 train-validation-test split, and CV+ conformal was applied with $k=20$ folds. Corresponding means and standard deviations appear in Table \ref{tab:cp_coverage_mean_std}.}
\begin{tabular}{llcccc}
\toprule
Sub-model & CP Method & $1-\alpha=90$\% & $1-\alpha=95$\% & $1-\alpha=98$\% & $1-\alpha=99$\% \\ 
\midrule
Reconstruction & Vanilla & 88.71 & 93.55 & 96.77 & 98.39 \\
Reconstruction & Split   & 87.10 & 93.55 & 95.97 & 97.58 \\
Reconstruction & CV+     & 89.52 & 94.35 & 97.58 & 98.39 \\
\midrule
Latent dynamics & Vanilla & 91.13 & 95.97 & 97.58 & 99.19 \\
Latent dynamics & Split   & 88.71 & 94.35 & 97.58 & 98.39 \\
Latent dynamics & CV+     & 95.97 & 99.19 & 100.00 & 100.00 \\
\midrule
End-to-end & Vanilla & 88.71 & 92.74 & 96.77 & 98.39 \\
End-to-end & Split   & 87.10 & 92.74 & 96.77 & 97.58 \\
End-to-end & CV+     & 91.13 & 95.16 & 97.58 & 99.19 \\
\bottomrule
\end{tabular}
\label{tab:cp_coverage_medians}
\end{table}

Across all CP methods, the empirical coverage was generally close to nominal levels, indicating good calibration. While split conformal is theoretically more accurate than vanilla conformal \cite{angelopoulos2021gentle,lei2018distribution}, for both reconstruction and end-to-end outputs, vanilla and split conformal performed similarly, with mean coverages typically within 1\%–2\% of the target and relatively low variability and standard deviations typically under $\sim$4\% (aside from when $1-\alpha=90\%$). By contrast, CV+ consistently achieved slightly higher accuracy, producing empirical coverages that were closer to the nominal rates in both mean and median, especially at higher confidence levels. This improvement was most apparent in the end-to-end model, where CV+ coverage levels tracked the nominal ones more tightly than vanilla or split conformal did.

The latent dynamical model displays somewhat different coverage behavior from what was observed for the reconstruction and end-to-end components. For vanilla and split CP, latent-space coverages are still near the nominal levels---typically within 1-2\%---but with slightly larger variability across samples than for reconstructions and end-to-end outputs. In contrast, CV+ produces systematically conservative intervals for the latent dynamics, with mean coverages around 95\% vs. 90\% at the 90\% nominal level and mean coverages exceeding 98\% at the 95–99\% nominal levels; in fact, the corresponding medians saturated at 100\% for the two highest nominal coverages. These results suggest that CP methods measure reconstruction and end-to-end prediction uncertainties reasonably well, while for latent dynamics, the coverage tends to be more conservative, especially for CV+, which provides the most stable yet also most pessimistic intervals.

To test more specifically for consistency in the predictive intervals across different CP methods, we also refer the reader to Figure \ref{fig:full_errors}. For all three subsets of the network, the predictive errors become less consistent as $\alpha\rightarrow 0$. This is because the convergence of empirical quantiles to their true values depends strongly on the miscoverage rate $\alpha$ \cite{serfling1980chapter2,vaart1998chapter21}. In particular, for CP, predictive intervals at smaller $\alpha$ require larger calibration sets to stabilize because extreme quantiles converge more slowly, whereas more central quantiles yield more stable intervals with fewer samples \cite{lei2018distribution,vovk_algorithmic_2022}. 

\section{Nonconformity scores}\label{tailwise}

In the more basic forms of conformal prediction, uncertainty intervals are constructed symmetrically---predictive errors above and below are assumed to have the same distribution. Concretely, for a model $f:X\rightarrow Y$, if we denote the \textbf{signed residual}
$$
R:=y-f(x)
$$
between the model prediction $f(x)$ for an input $x\in X$ and a true outcome $y\in Y$, then the \textit{absolute} residuals $|R|$ are often used to calibrate a single quantile. Thus, the prediction interval is symmetric and can be written as 
$$
\left[f(x)-Q_{1-\alpha}(|R|),f(x)+Q_{1-\alpha}\left(|R|\right)\right],
$$
where $Q_{1-\alpha}\left(|R|\right)$ is the $(1-\alpha)$-quantile of $|R|$ as computed empirically over the dataset.

The aforementioned construction is simple and guarantees the desired coverage over the calibration data, but it forces the lower and upper bounds to be equally wide. Hence, following the tailwise quantile approach first introduced in \cite{barber_predictive_2021}, we use tailwise (one-sided) quantiles for the \textbf{DSD-valued} predictions. For a desired miscoverage $\alpha$, we split the miscoverage evenly between the tails on both sides, with $\alpha/2$ mass for each. We then compute the lower and upper quantiles for the \textit{signed} residuals: $Q_{\alpha/2}(R)$ and $Q_{1-\alpha/2}(R)$. This defines the prediction intervals used in this study:
$$
\left[
f(x)+Q_{\alpha/2}(R),f(x)+Q_{1-\alpha/2}(R)
\right],
$$
which still ensures $1-\alpha$ coverage over the validation set but allows the lower and upper margins to differ whenever the residual distribution is asymmetric.

For \textbf{latent-space predictions}, the outputs are multivariate and typically exhibit correlations across dimensions. Hence, in this case, coordinate-wise absolute or signed residuals may not accurately capture the joint error structure. Accordingly, for a predicted latent vector $\hat{z}=f(x)$ and true latent vector $z$, we define the residual $r = z - \hat{z}$ and use a scalar nonconformity score based on the (squared) \textbf{Mahalanobis distance}:
$$
S(z,\hat{z})
= r^\top \Sigma_r^{-1} r,
$$
where $\Sigma_r$ is the empirical covariance matrix of latent-space residuals estimated from the calibration set (in practice, using a Ledoit–Wolf shrinkage estimator \citep{LedoitWolf2012,scikit-learn}). In particular, the scalar score $S(z,\hat{z})$ is used to compute the empirical $(1-\alpha)$-quantile on the calibration set, which then defines the size of the latent-space prediction ellipsoid. Using the \textit{residual} covariance rather than the latent covariance yields a conformity score that is properly normalized with respect to the model's error geometry and captures correlated uncertainty through multivariate prediction ellipsoids.

The prediction intervals and nonconformity scores described above are computed independently at each timestep, using instantaneous residuals and covariance estimates. Hence, the prediction sets evolve in time, therby reflecting the temporal evolution of model uncertainty.

\section{Reproducibility Details for AE-SINDy Model Training}\label{reproducibility}

This appendix provides the essential information required to reproduce the autoencoder-SINDy (AE-SINDy) model setup and training procedure used in this study. The description covers data acquisition and preprocessing, model architecture and hyperparameters, and the training workflow, including loss function specification. The source code and data are further included in a linked repository. Further details on the AE-SINDy architecture, the polynomial SINDy library, and the loss structure used during training are provided in the companion study De Jong et al. \cite{dejong2025data}, which develops the surrogate model in full.

\subsection{Data Source and Preprocessing}\label{data}

The AE-SINDy model is trained using binned particle size distribution (PSD) data generated from a large-eddy simulation (LES) employing the superdroplet method. We simulate the evolution of a warm liquid-phase cloud that forms from a Gaussian surface moisture and heat flux, growing in altitude before precipitating. The primary datasets are accessed in NetCDF format and contain variables for binned droplet mass distributions over time and space, with droplet coalescence active as the only enabled droplet dynamic. For model input, only samples with sufficient liquid water content (e.g., exceeding $10^{-5}$~kg/kg) are included. Each PSD is normalized by its total liquid mass to ensure scale invariance during encoding. 

The dataset is partitioned into training (80\%, 494 samples) and testing (20\%, 124 samples) sets. For each sample, the normalized PSD and its time derivative (computed via finite differences) are paired with the corresponding total liquid mass. The input tensors are shaped as $(N_\text{batch}, N_\text{bins})$ for both the PSD and its time derivative, where $N_\text{bins} = 64$. The total mass is provided as an additional input feature that bypasses the encoder to become the final latent variable. Total mass is further rescaled during training and testing by the maximum value of total mass contained in the training dataset.

For a full description of the large eddy simulations used to generate the PSD training data---including the simulation setups, temporal sampling, grid resolution, and the physics assumed in the forward simulation---see De Jong et al. \cite{dejong2025data}, which details the simulation design and post-processing used to produce the binned PSD datasets.

\subsection{Model Architecture and Hyperparameters}

The AE-SINDy architecture used here follows the structure introduced in \cite{dejong2025data}, which employs four fully-connected encoder and decoder layers that halve (or double, respectively) the dimension at each layer, using ReLU activations in the hidden layers and a softmax output to ensure normalized reconstructions, and employs a SINDy model for the dynamics in the latent space. More explicitly, the model consists of three components:

\begin{enumerate}
\item \textbf{Encoder:} A feed-forward neural network (FFNN) with four fully connected layers that sequentially reduce the input dimension from $N_\text{bins}$ to the latent dimension (excluding the total-mass variable), yielding $N_\text{latent} - 1 = 3$ latent variables. Hidden layers use ReLU activations, and the final layer maps to the latent space without a nonlinearity.
\item \textbf{Decoder:} A FFNN mirroring the encoder structure, with four fully connected layers expanding from $N_\text{latent} - 1$ back to $N_\text{bins}$. A softmax activation on the output layer enforces that reconstructed PSDs remain normalized.
\item \textbf{SINDy Dynamics Module:} A bias-free, single-layer neural network that implements the SINDy formulation by outputting a linear combination of latent-space time derivatives using a polynomial feature library of terms up to second order.
\end{enumerate}

Key hyperparameters for the model are as follows:

\begin{itemize}
    \item \texttt{latent\_dim}: Number of latent variables (4 total: 3 from the PSD encoding, plus 1 for the total liquid mass)
    \item \texttt{poly\_order}: Maximum polynomial order in SINDy library (2)
    \item \texttt{batch\_size}: Training batch size (25)
    \item \texttt{learning\_rate}: Initial learning rate for AdamW optimizer (e.g., 0.0042)
    \item \texttt{patience}: Early stopping patience (50 epochs)
    \item \texttt{weight\_decay}: L2 regularization coefficient ($10^{-3}$)
    \item \texttt{tol}: Numerical tolerance for loss calculations ($10^{-8}$)
    \item \texttt{loss\_weights}: Relative weights for loss terms, determined via Champion et al.'s recommended scaling (see code for details)
\end{itemize}

All network weights are initialized using Xavier or Kaiming normal initialization, with zero bias.

\subsection{Training Procedure and Loss Function}

Training procedures for the AE-SINDy architecture follow the approach detailed in \cite{dejong2025data} and the associated code repository, including early stopping, learning-rate scheduling, and a composite loss combining reconstruction, PSD-derivative, and latent-derivative terms.

In summary, training is performed using the AdamW optimizer with learning rate scheduling and early stopping based on validation loss. The model is trained for up to 1000 epochs, with the option to halt training if no improvement is observed over a specified patience interval.

The total loss function $L$ is a weighted sum of three components:

\begin{align}
    L = L_\text{recon} + w_{dx} L_{dx} + w_{dz} L_{dz}
\end{align}

where:
\begin{itemize}
    \item $L_\text{recon}$ is the Kullback-Leibler divergence between the normalized input PSD and its reconstruction.
    \item $L_{dx}$ is the mean squared error between the predicted and actual time derivative of the PSD, projected via the decoder.
    \item $L_{dz}$ is the mean squared error between the predicted and actual time derivative in the latent space, as computed by the SINDy module.
\end{itemize}

Loss weights are chosen to balance the reconstruction and dynamics learning, following a scaling based on the relative magnitudes of the PSD and its time derivative in the training data as in Champion et al. \cite{champion_data-driven_2019}. Other parameters, including the batch size, initial learning rate, and a multiplicative factor of $w_{dx}$, were determined using hyperparameter optimization with Optuna \cite{Optuna}.

\subsection{Code Availability}

All code used for data processing, model definition, and training is written in Python using PyTorch and is available at \url{https://github.com/jonaskat87/UQ_AE-SINDy}. The scripts include utilities for loading NetCDF datasets, constructing PyTorch DataLoaders, defining the AE-SINDy architecture, executing the training loop, and running and visualizing the uncertainty quantification pipelines.

Additional scripts for data processing, model specification, and training of the AE-SINDy surrogate are available in the code repository accompanying the companion paper De Jong et al. \cite{dejong2025data}.

\end{document}